\documentclass{bmvc2k}

\newcommand\blfootnote[1]{
  \begingroup
  \renewcommand\thefootnote{}\footnote{#1}
  \addtocounter{footnote}{-1}
  \endgroup
}

\title{\textbf{ISG}: I can See Your Gene Expression}

\addauthor{Yan Yang}{u6169130@anu.edu.au}{1}
\addauthor{LiYuan Pan}{liyuan.pan@bit.edu.cn}{2}
\addauthor{Liu Liu}{nwpuliuliu@gmail.com}{3}
\addauthor{Eric A Stone}{eric.stone@anu.edu.au}{1$^{\dagger}$}

\addinstitution{
 Biological Data Science Institute, \\
 Research School of Biology, \\
 The Australian National University,\\
 Australia
}

\addinstitution{BITSZ \& School of CSAT, BIT, China}
\addinstitution{Cyberverse Lab, China}

\runninghead{Yang, Pan, Liu, Stone}{ISG: I can See Your Gene Expression}

\usepackage{lipsum}
\usepackage{amsmath}
\usepackage{amsfonts}
\usepackage{amssymb}
\usepackage{cleveref}
\crefname{section}{Sec.}{Secs.}
\Crefname{section}{Section}{Sections}
\Crefname{table}{Table}{Tables}
\crefname{table}{Tab.}{Tabs.}
\Crefname{equation}{Equation}{Equations}
\crefname{equation}{Eq.}{Eqs.}
\Crefname{algorithm}{Algorithm}{Algorithms}
\crefname{algorithm}{Alg.}{Algs.}
\Crefname{figure}{Figure}{Figure}
\crefname{figure}{Fig.}{Fig.}
\Crefname{appendix}{Appendix}{Appendix}
\crefname{appendix}{Appx.}{Appx.}

\newcommand{\name}{\textbf{ISG}}
\newcommand{\extractor}{\textit{Feature Extraction}~}

\usepackage{enumitem}
\setlist[enumerate]{itemsep=0mm}

\usepackage{booktabs}
\usepackage{multirow}
\usepackage{wrapfig}
\usepackage{cuted}
\definecolor{cadmiumgreen}{rgb}{0.0, 0.42, 0.24}

\allowdisplaybreaks

\definecolor{brightmaroon}{rgb}{0.76, 0.13, 0.28}

\begin{document}

\maketitle
\blfootnote{$^{\dagger}$ Corresponding author.}
\begin{abstract}
This paper aims to predict gene expression from a  histology slide image precisely. Such a slide image has a large resolution and sparsely distributed textures. These obstruct extracting and interpreting discriminative features from the slide image for diverse gene types prediction. Existing gene expression methods mainly use general components to filter textureless regions, extract features, and aggregate features uniformly across regions. However, they ignore gaps and interactions between different image regions and are therefore inferior in the gene expression task (\cref{sec:intro}).  
Instead, we present \name~framework that harnesses interactions among discriminative features from texture-abundant regions by three new modules: 1) a \textit{Shannon Selection} module (\cref{sec:shannon_selection}), based on the Shannon information content and Solomonoff's theory, to filter out textureless image regions;
2) a \extractor network (\cref{sec:compression}) to extract expressive low-dimensional feature representations for efficient region interactions among a high-resolution image; 3) a \textit{Dual Attention} network (\cref{sec:network}) attends to regions with desired gene expression features and aggregates them for the prediction task. Extensive experiments on standard benchmark datasets show that the proposed \name~framework outperforms state-of-the-art methods significantly.
\end{abstract}

\section{Introduction}
\label{sec:intro}

Gene expression prediction from a histology slide image is an indispensable component for efficiently understanding clinic treatment developments \cite{rna-seq,stnet,stain_learning}.  
A histology slide image has two characteristics: i) it has a large resolution amounting to $10^{5} \times 10^{5}$ \cite{rna-seq}. The large resolution prevents an end-to-end solution, i.e., directly using traditional deep learning approaches (e.g., convolution \cite{resnet} and transformer \cite{att,vit} networks), for high computational cost; 
ii) it has sparse and non-uniformly distributed textures which hinder model inference \cite{vitpd}. To predict gene expression precisely, feature extractions and interactions of regions with different texture levels among the histology slide image need to be explored.

\begin{figure}[!t]
    \centering
    \includegraphics[width=\linewidth]{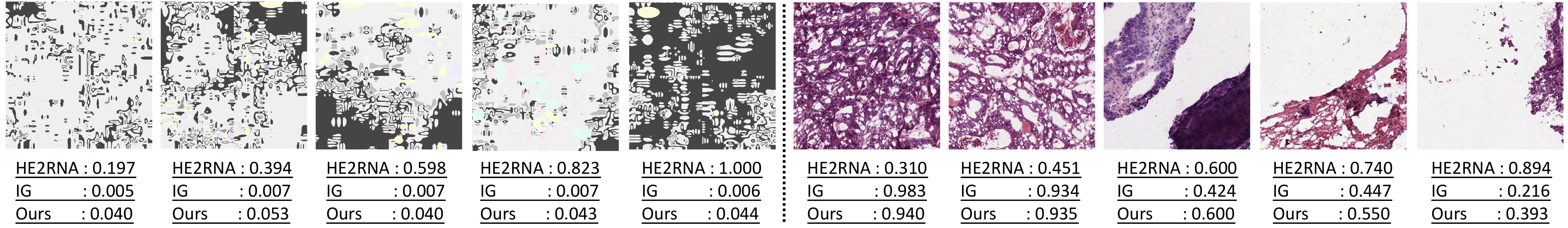}
    \caption{\it  
    Comparison of the proposed Shannon Selection module with HE2RNA \cite{rna-seq} and the conventional image gradient (IG) based selection method. \textbf{Columns} $1^{st}$-$5^{th}$: textureless patches. Note that HE2RNA fails to assign  consistent scores for these patches and the scores are high. Both IG and our Shannon Selection module assign consistent low scores for these patches.   \textbf{Columns} $6^{th}$-$10^{th}$: patches with various degree of textures/features. Though both IG and our Shannon Selection module assign high scores for these patches, they prioritize texture/feature abundance differently (e.g., $8^{th}$ and $9^{th}$ columns), resulting in different filtered patches and prediction accuracies (See comparison in \cref{tab:selection_compare}).}
    \label{fig:feature_selection}
\end{figure}

To date, this gene prediction problem remains under-explored. The pioneer HE2RNA \cite{rna-seq} provides a three-stage solution. First, it tiles a histology slide image into patches and filters out patches with high background noise via the Otsu algorithm \cite{otsu}. 
Second, patch features are extracted from a pretrained ResNet \cite{resnet} and clustered with a K-mean algorithm by patch locations.
Third, the aggregated cluster-wise features are independently processed by a multi-layer perceptron (MLP) \cite{mlp} and an average pooling layer, for gene expression prediction. However, HE2RNA has three limitations: 
i) Textureless patches, e.g., patches with a solid colour or scattered noisy chunks (see \cref{fig:feature_selection}), are not filtered;
ii) The ResNet pre-trained on ImageNet fails to identify histology-related features due to dataset gaps (see \cref{sec:abl}); 
and iii) feature interactions among patches are not considered, which neglects long-range dependency between patches for gene expression predictions.

\begin{figure}[!t]
    \centering
    \includegraphics[width=\linewidth]{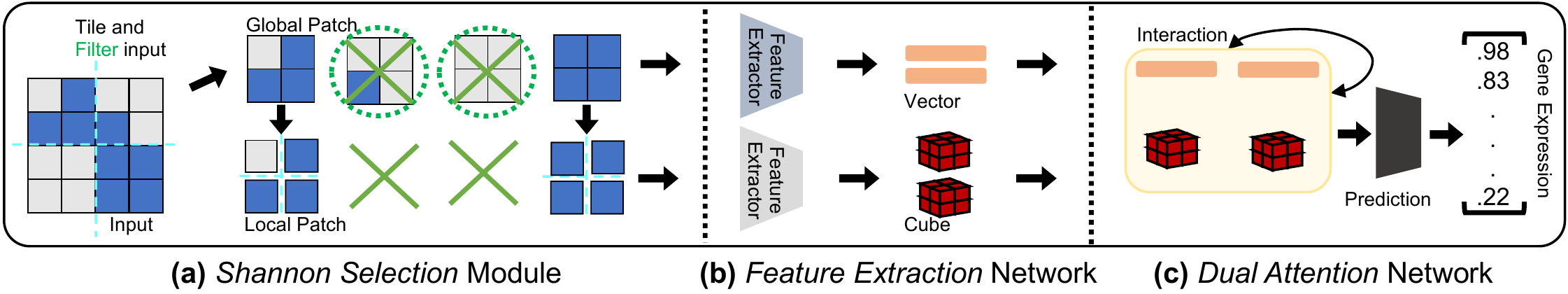}
    \caption{\it \name~framework. \textbf{(a)} \textcolor{blue}{Blue} and \textcolor{gray}{gray} squares separately denote texture-abundant and textureless patches. Given an input image, it is tiled using two (coarse and fine) resolutions, resulting in global and local patches. Textureless/featureless patches are  \textcolor{cadmiumgreen}{filtered} out with a \textit{Shannon Selection} module.  
    \textbf{(b)} Taking global and local patches as input, two separate \extractor networks are used to extract  low-dimension feature representations, resulting in  \textcolor{orange}{global} and \textcolor{brightmaroon}{local} features. Each \textcolor{orange}{global} feature corresponds with multiple \textcolor{brightmaroon}{local} features, or equivalently, a \textcolor{brightmaroon}{local} feature cube. \textbf{(c)} Taking global and local features as input, our \textit{Dual Attention Network} brings interactions to these two types of features before predictions.
    }
    \label{fig:intro}
\end{figure}

To address the above limitations, by analysing the characteristics of the histology slide image, we propose an \name~framework (see \cref{fig:intro}) that builds feature interactions between patches with abundant textures and injects global contextual information into features to make better predictions. Our \name~has three new modules connected in a sequence: 1) a theoretical \textit{Shannon Selection} module (\cref{sec:shannon_selection}). It quantifies the patch texture abundance levels. Given an input histology slide image, it is first segmented into patches at two resolutions - coarse and fine. We separately name the `coarse-resolution' and `fine-resolution' patches to `global' and `local' patches. The \textit{Shannon Selection} module selects patches with a large length of minimal description by incorporating Solomonoff's universal prior and Shannon information content; 2) a \extractor network (\cref{sec:compression}). 
Given  `global' and `local' patches, two separate \extractor networks are used to extract discriminative patch representations. 
Both fine-grained local features and coarse-level global features are obtained. This module follows an unsupervised manner, as there is a relatively large amount of images compared to the available label annotations (e.g., each gene expression label pairs with an image with up to $10^{5} \times 10^{5}$; and 3) a \textit{Dual Attention} network (\cref{sec:network}). It takes global and local features as inputs, brings interactions to them, and predicts gene expression. 
  
Our contributions are summarized as follows: 
1) a new \name~framework is proposed to predict gene expression from a histology slide image; 
2) a new theoretical \textit{Shannon Selection} module is proposed to filter out textureless image patches; 
3) a new \extractor network is proposed to extract discriminative patch features in an unsupervised manner;
4) a new \textit{Dual Attention} network is proposed to calibrate patch features by injecting global contextual information to features and make better predictions; 
and 5) Our model outperforms state-of-the-arts (SOTA) methods significantly.

\section{Related Works}
\noindent\textbf{Computational Biomedical Domain.} 
Deep learning demonstrates significant milestones in assisting disease diagnosis including cancer classification \cite{c1,c2}, biomedical image segmentation \cite{seg1,seg2}, tumor mutational burden prediction \cite{m1,m2}, and gene expression prediction \cite{rna-seq,stain_learning,stnet}. Gene expression prediction is the most essential and attractive task that would facilitate understanding and designing novel treatments \cite{stnet}. Two sub-problems have been derived for two distinct gene expression profiling techniques \cite{bulk,spatial}. First, Schmauch \textit{et al.} present a HE2RNA to model bulk RNA-Seq \cite{bulk}. It targets quantifying gene expression for a whole histology sample which is up to $10^{5} \times 10^{5}$ resolution {(as a reference, this resolution is larger than the majority of remote sensing images \cite{remote})}. They introduce a three-stage solution, including K-means and transfer learning, to extract image-level features. However, the model performance is promising to be further improved with a more task-specific design. Second, Dawood \textit{et al.} \cite{stain_learning}, He \textit{et al.} \cite{stnet}, and Zeng \textit{et al.} \cite{st_transformer} introduce NSL, STNet, and Hist2ST to measure spot-level gene expression of a histology slide image from a spatial transcriptomics (ST)-based \cite{spatial} dataset. The ST technique is still under development, and there remains a wide audience to bulk RNA-Seq. Meanwhile, existing ST datasets lack diversities that typically contain tens of patients \cite{stnet}. Thus, this paper follows HE2RNA \cite{rna-seq} to model bulk RNA-Seq from histology slide images.

\noindent\textbf{Representational Learning.} There has been a huge effort from computer vision community on studying unsupervised feature learning \cite{utl1,sia1,utl2,sia2,cor1}. 
With data augmentation crafted image views, a contrastive learning framework \cite{utl1,sia1,sia2} usually learns a similarity-based representation from positive and negative matched view pairs.
Alternatively, it enforces a unit cross-correlation matrix that is calculated from embeddings of two views of the same image \cite{cor1}. 
However, these methods are hard to train because of requiring a large batch size and occasionally confront model collapses \cite{utl2}. Instead, a StyleGAN \cite{stylegan2,stylegan3} has high accessibility, and it delivers a low dimension and versatile feature representation of a high-resolution image \cite{stylespace,psp,s2f}. This is inevitable before establishing interactions of regions/patches among a histology slide image. In this paper, we investigate the use of StyleGAN for pre-learning gene expression features.

\section{Methodology}

\noindent \textbf{Problem Formulation.} 
Given a histology slide image $\mathbf{X} \in \mathbb{R}^{\mathsf{h} \times \mathsf{w} \times 3}$, we aim to predict its associated gene expression $\mathbf{Y} \in \mathbb{R}^{\mathsf{n} \times 1}$, where $\mathsf{h}$, $\mathsf{w}$, and $\mathsf{n}$ are height, width, and number of gene types, respectively. Our framework is given in \cref{fig:intro}.

We first tile $\mathbf{X}$ into patches at a coarse-resolution and build a patch set $\mathcal{X}^{G} = \{\mathbf{x}_{i} \in \mathbb{R}^{\mathsf{p} \times \mathsf{p} \times 3} \mid i \in 1,\cdots, \lfloor \frac{\mathsf{h}\times \mathsf{w}}{\mathsf{p}^{2}} \rfloor\}$, where $\mathsf{p}$ is the coarse patch size and $\lfloor \cdot \rfloor$ denotes the $\mathsf{floor}$ operator. $\mathcal{X}^{G}$ is further fed to a selector $\mathbf{S}(\cdot)$ to filter out textureless patches, resulting in a subset $\ddot{\mathcal{X}}^{G} = \mathbf{S}(\mathcal{X}^{G})$ (\cref{sec:shannon_selection}). 
For each $\mathbf{x}_{i} \in \ddot{\mathcal{X}}^{G}$, we tile it into fine-grained patches at a fine-resolution, resulting in $\ddot{\mathcal{X}}^{L}_{i} = \{\mathbf{x}_{i,j} \in \mathbb{R}^{\mathsf{q} \times \mathsf{q} \times 3} \mid j \in 1,\cdots,\lfloor \frac{\mathsf{p}^{2}}{\mathsf{q}^{2}} \rfloor\}$, where $\mathsf{q}$ is the fine patch size and $\mathsf{q} < \mathsf{p}$. Collecting all fine patches yields a fine patch set $\ddot{\mathcal{X}}^{L}$.

Given $\ddot{\mathcal{X}}^{G}$ and $\ddot{\mathcal{X}}^{L}$, we separately use two feature extractors $\mathbf{E}^{G}(\cdot)$ and $\mathbf{E}^{L}(\cdot)$ to extract patch-wise low-dimension feature representation (\cref{sec:compression}). For each patch $\mathbf{x}_{i} \in \ddot{\mathcal{X}}^{G}$, we have $\mathbf{f}_{i} = \mathbf{E}^{G}(\mathbf{x}_{i})$ and $\mathbf{f}_{i} \in \mathbb{R}^{\mathsf{d} \times 1}$, where $\mathsf{d}$ is the feature dimension. Collecting all coarse patch feature vectors yields a global feature set $\mathcal{F}^{G} = \{\mathbf{f}_{i} \mid i \in 1,\cdots,|\ddot{\mathcal{X}}^{G}|\}$.  For each fine-grained patch set $\ddot{\mathcal{X}}^{L}_{i} \in \ddot{\mathcal{X}}^{L}$, it is corresponding to the patch $\mathbf{x}_{i}$. We extract patch-wise feature vectors for each fine-grained patch, arrange them according to their relative positions within $\mathbf{x}_{i}$ and obtain a feature map  $ \mathbf{R}_{i} = \mathbf{E}^{L}(\ddot{\mathcal{X}}^{L}_{i})$, where $ \mathbf{R}_{i} \in \mathbb{R}^{\frac{\mathsf{p}}{\mathsf{q}} \times \frac{\mathsf{p}}{\mathsf{q}} \times \mathsf{d}}$. Collecting all fine patch feature maps yields a local feature set $\mathcal{F}^{L} = \{\mathbf{R}_{i} \mid i \in 1,\cdots,|\ddot{\mathcal{X}}^{G}|\}$.

With $\mathcal{F}^{G}$ and $\mathcal{F}^{L}$, a dual attention network $\mathbf{M}(\cdot,\cdot)$ (\cref{sec:network}) is proposed to fuse them and predicts the gene expression  $\mathbf{Y} = \mathbf{M}(\mathcal{F}^{G},\mathcal{F}^{L})$.

\subsection{Patch Selection Module}
\label{sec:shannon_selection}

A histology slide image has large textureless/featureless regions. It is natural to select feature abundant patches by using edges, as edges are units to form features \cite{edge}. A number of edge detectors are available, e.g., Image Gradient (IG), Canny Edge Detector (CANNY), DexiNed (Dex) \cite{dex}, Difference of Gaussian (DoG), and Laplacian of Gaussian (LoG). Instead of using edges, in this work, we give an alternative view from information theory to identify feature abundant patches. The comparison of our method and edge detection-based patch selectors are given in \cref{sec:exp}.

\noindent \textbf{Shannon selection.}
For a patch $\mathbf{x}_{i}$, we use the Shannon information content \cite{entropy} to quantify its texture/feature abundance level. The quantity is given by ${h}(\mathbf{x}_{i}) = \log_2 \frac{1}{\Pr(\mathbf{x}_{i})}$, where $\Pr(\cdot)$ is the probability mass function of  $\mathbf{x}_{i}$. Note,  ${h}(\mathbf{x}_{i})$ measures bit quantities of a patch. For a patch with poor features, ${h}(\mathbf{x}_{i}) \rightarrow 0$.

The key is to find a $\Pr(\cdot)$ that describes $\mathbf{x}_{i}$ well. Following \cite{mdl,uai}, we employ Solomonoff’s universal prior $\Pr(\mathbf{x}_{i}) = \sum_{p : \mathcal{U}(p) = \mathbf{x}_{i} * } 2^{-\lVert p \rVert_{0}}$ as our patch distribution, where $\lVert \cdot \rVert_0$ is the length calculator, $p$ is a program fed into a universal Turing machine $\mathcal{U}(\cdot)$, and $*$ denotes any possible suffix. This prior considers each feasible program $p$ that derives `$\mathbf{x}_{i} *$', i.e., any string starts with a bit representation of $\mathbf{x}_{i}$, from $\mathcal{U}(\cdot)$. Afterwards, it sums over the negative exponent of the program length.

\vspace*{-2mm}
\begin{equation}
    h(\mathbf{x}_{i}) = -\log \sum_{p : \mathcal{U}(p) = \mathbf{x}_{i} * } 2^{-\lVert p \rVert_{0}} \approx - \log 2^{-K(\mathbf{x}_{i})} = K(\mathbf{x}_{i}), \label{eq:shannon} 
\end{equation}
where $K(\cdot)$ is the Kolmogorov complexity, and $K(\mathbf{x}_{i})$ is the shortest program length for the input $\mathbf{x}_{i}$ and an excellent approximation of the Solomonoff's prior \cite{uai}.
\cref{eq:shannon} suggests that a patch $\mathbf{x}_{i}$ deriving a large bit quantity tends to have abundant features. With a preset threshold, we select patches with abundant features.

\begin{figure}[!t]
    \centering
    \includegraphics[width=\linewidth]{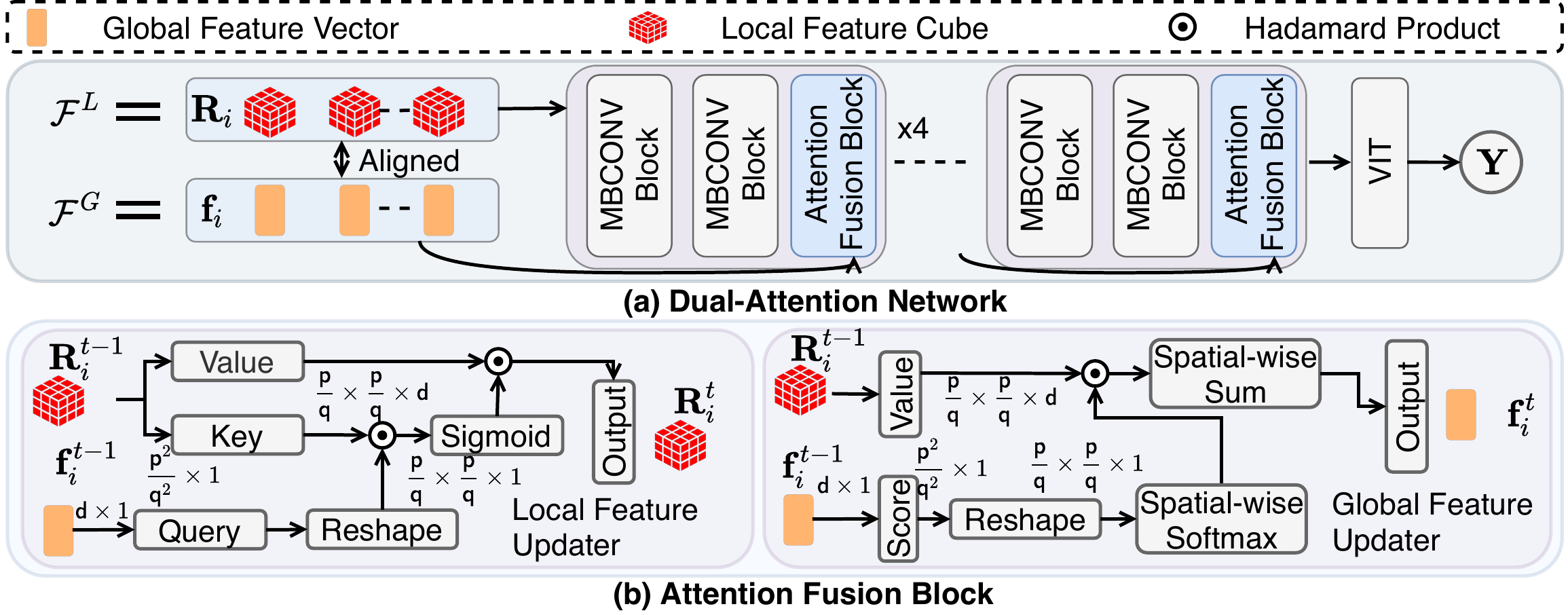}
    \caption{\it The architecture of our \textit{Dual Attention} network. \textbf{(a)} 
    For each $\mathbf{R}_{i}$ and $\mathbf{f}_{i}$ pair, we refine them with MBCONV \cite{efficientnet} and attention fusion blocks, followed by a small ViT \cite{vit} for predicting the gene expression. 
    \textbf{(b)} An attention fusion block consists of a local feature updater (Left) and a global feature updater (Right). A local feature updater uses the global feature vector $\mathbf{f}_{i}$ as the guidance, updating local features to emphasize features from regions of interest. A global feature updater uses the local feature $\mathbf{R}_{i}$ as the guidance, updating global features to reflect the evolving significance of local features. }  
    \label{fig:arch}
\end{figure}

\subsection{Feature Extraction Network}
\label{sec:compression}

For selected patches, we extract low-dimensional patch-wise features in this section. 
We use a style-based architecture for our feature extractor as it can capture versatile feature representations in an unsupervised manner \cite{psp,stylespace,s2f}.  Here, our extractor is from training a styleGAN-based autoencoder with image reconstruction as an auxiliary task.

\noindent \textbf{Method.}
Let $\mathbf{D}(\cdot)$ be a styleGAN (aka. decoder) \cite{stylegan2} and $\mathbf{E}(\cdot)$ be an associated extractor (aka. encoder) \cite{s2f}. We use two different groups of them, $\mathbf{D}^{G}(\mathbf{E}^{G}(\cdot))$ and $\mathbf{D}^{L}(\mathbf{E}^{L}(\cdot))$, to learn salient features $\mathcal{F}^{G}$ and $\mathcal{F}^{L}$ from global and local patches, respectively. When extracting coarse-level features, $\mathbf{E}^{G}(\cdot)$ takes global patches as inputs. When extracting fine-grained features, $\mathbf{E}^{L}(\cdot)$ takes local patches as inputs. We empirically verify the versatility and effectiveness of our global and local feature extractors in \cref{sec:exp}.     

\noindent \textbf{Objectives.} 
For brevity, we omit the superscript of $\mathbf{D}(\cdot)$ and $\mathbf{E}(\cdot)$, as the global and local feature extractors are trained in the same manner. 
To train them, we use the  $\mathcal{L}_{1}$ loss, the AlexNet-based LPIPS loss $\mathcal{L}_{\text{LPIPS}}$ \cite{alexnet,lpips}, and the discriminator loss $\mathcal{L}_{C}$ (with a discriminator $\mathbf{C}(\cdot)$) \cite{stylegan2}, where $\mathcal{L}_{1} = \lVert \mathbf{x} - \mathbf{D}(\mathbf{E}(\mathbf{x})) \rVert$, $\mathcal{L}_{\text{LPIPS}} = \lVert \phi(\mathbf{x}) - \phi(\mathbf{D}(\mathbf{E}(\mathbf{x}))) \rVert_{2}$, and  $ \mathcal{L}_{C} = u\big(\mathbf{C}(\mathbf{x})\big) + u\big(-\mathbf{C}(\mathbf{D}(\mathbf{E}(\mathbf{x})))\big)$. $u(\cdot)$ is the Softplus function and $\phi(\cdot)$ is a pretrained LPIPS network.

The $\mathcal{L}_{1}$ loss and the LPIPS loss ensure our image reconstruction fidelity while affecting feature extraction quality \cite{psp,s2f}. The discriminator loss is an indispensable objective function for style-based architecture. Our final training objective is given by 

\vspace*{-7mm}
\begin{align}
    \mathcal{L}_{\text{total}} = \min_{\mathbf{D},\mathbf{E}} \max_{\mathbf{C}} ~ \mathbb{E}_{\mathbf{x} \sim \mathbf{X}} \big[ \mathcal{L}_{1} + \mathcal{L}_{\text{LPIPS}} + \mathcal{L}_{C}\big].
\end{align}

\subsection{Dual Attention Network}
\label{sec:network}

With global features $\mathcal{F}^{G} = \{\mathbf{f}_{i} \mid i \in 1,\cdots,|\ddot{\mathcal{X}}^{G}|\}$ and local features $\mathcal{F}^{L} = \{\mathbf{R}_{i} \mid i \in 1,\cdots,|\ddot{\mathcal{X}}^{G}|\}$,
we propose a \textit{Dual Attention} network to adaptively calibrate model attention to regions of interest.

\noindent \textbf{Method.} 

The architecture of our \textit{Dual Attention} network is given in \cref{fig:arch}. It has two modules connected in a sequence:  1) a lightweight MBCONV block sequence \cite{efficientnet} interleaved with attention fusion blocks to jointly refine the local feature cube $\mathbf{R}_{i}$ and global feature $\mathbf{f}_{i}$; and 2) a small vision transformer \cite{vit} take  average-pooled  local feature cubes 
as input and predict the gene expression. Our attention fusion block has two components, the local feature updater and the global feature updater.  
\noindent{\it (a) Local Feature Updater.} Let $t$ be the layer index. Each feature vector in the local feature cube $\mathbf{R}^{t-1}_{i}$ represents a local patch, while having different priorities for predicting the gene expression. With the guidance from the global feature $\mathbf{f}^{t-1}_{i}$, we calibrate the local feature cube $\mathbf{R}^{t-1}_{i}$.

Specifically, we first project the global feature vector  $\mathbf{f}^{t-1}_{i} \in \mathbb{R}^{\mathsf{d} \times 1}$ to obtain the query $\mathbf{Q}_{i}^{t}$ and the local feature cube $\mathbf{R}^{t-1}_{i} \in \mathbb{R}^{\frac{\mathsf{p}}{\mathsf{q}} \times \frac{\mathsf{p}}{\mathsf{q}} \times \mathsf{d}}$ to obtain the key $\mathbf{K}_{i}^{t}$ and the value $\mathbf{V}_{i}^{t}$. 

\vspace*{-5mm}
\begin{align}
    \mathbf{Q}^{t}_{i} = \text{Reshape} (\mathbf{W}^{t}_{q}\mathbf{f}^{t-1}_{i}), \quad \quad 
    \mathbf{K}^{t}_{i} = \mathbf{R}^{t-1}_{i} \mathbf{W}^{t}_{k}, \quad \quad 
    \mathbf{V}^{t}_{i} =  \mathbf{R}^{t-1}_{i} \mathbf{W}^{t}_{v}, 
\end{align}

\vspace*{-0.7em}
\noindent where $\mathbf{W}^{t}_{q} \in \mathbb{R}^{ \frac{\mathsf{p}^{2}}{\mathsf{q}^{2}} \times \mathsf{d}},\mathbf{W}^{t}_{k} \in \mathbb{R}^{\mathsf{d} \times \mathsf{d}}$, and $\mathbf{W}^{t}_{v} \in \mathbb{R}^{\mathsf{d} \times \mathsf{d}}$ are weight matrices.

We then compute the correlation of the query and key, followed by a Sigmoid activation function, to obtain a score map $\mathbf{A}^{t}_{i}$. Finally, the score map $\mathbf{A}^{t}_{i}$ is used to modulate the value $\mathbf{V}_{i}^{t}$ to obtain the updated local feature cube $\mathbf{R}^{t}_{i}$.

\vspace*{-5mm}
\begin{align}
    \mathbf{A}^{t}_{i} &= \text{Sigmoid} \big(\mathbf{Q}^{t}_{i} \odot \mathbf{K}^{t}_{i}\big), \quad&& \mathbf{Q}^{t}_{i} \in \mathbb{R}^{\frac{\mathsf{p}}{\mathsf{q}} \times \frac{\mathsf{p}}{\mathsf{q}} \times 1}, ~ \mathbf{K}^{t}_{i} \in \mathbb{R}^{\frac{\mathsf{p}}{\mathsf{q}} \times \frac{\mathsf{p}}{\mathsf{q}} \times \mathsf{d}}, \label{eq:local_attention} \\
    \mathbf{R}^{t}_{i} &= \big( \mathbf{A}^{t}_{i} \odot \mathbf{V}^{t}_{i} \big) \mathbf{W}^{t}_{r} , \quad&& \mathbf{A}^{t}_{i}, ~ \mathbf{V}^{t}_{i} \in \mathbb{R}^{ \frac{\mathsf{p}}{\mathsf{q}} \times \frac{\mathsf{p}}{\mathsf{q}} \times \mathsf{d}}, ~  \mathbf{W}^{t}_{r} \in \mathbb{R}^{\mathsf{d} \times \mathsf{d}}. 
\end{align}
Here, $\mathbf{W}^{t}_{r}$ is an weight matrix, $\odot$ and $\text{Sigmoid}(\cdot)$ are the Hadamard product and the Sigmoid function. Unlike vanilla  attention mechanism \cite{transformer}, we directly scale the value $\mathbf{V}_{i}^{t}$ by the score map $\mathbf{A}^{t}_{i}$,  instead of costly aggregating it with matrix multiplications.

\noindent{\it (b) Global Feature Updater.} 
With the guidance from the local feature cube $\mathbf{R}^{t-1}_{i}$, we calibrate the global feature vector $\mathbf{f}^{t-1}_{i}$ to reflect the evolving significance of local features. Specifically, we first project the global feature vector $\mathbf{f}^{t-1}_{i}$ to the same spatial dimension with $\mathbf{R}^{t-1}_{i}$, and obtain a score matrix. We then normalize the score matrix spatially with a Softmax activation function to obtain a weight map $\mathbf{Z}^{t}_{i}$.  Finally, we compute the Hadamard product between the weight map $\mathbf{Z}^{t}_{i}$ and projected local feature $\mathbf{P}^{t}_{i}$, followed by a sum-aggregation to obtain the updated global feature vector $\mathbf{f}^{t}_{i}$. Mathematically, we have

\vspace*{-5mm}
\begin{align}
    \mathbf{Z}^{t}_{i} &= \text{Softmax} \big( \text{Reshape} ( \mathbf{W}^{t}_{z} \mathbf{f}^{t-1}_{i}) \big), \quad&&\mathbf{W}^{t}_{z} \in \mathbb{R}^{\lfloor \frac{\mathsf{p}^{2}}{\mathsf{q}^{2}} \rfloor \times \mathsf{d}},~\mathbf{f}^{t-1}_{i} \in \mathbb{R}^{\mathsf{d} \times 1}, \\
    \mathbf{P}^{t}_{i} &= \mathbf{R}^{t-1}_{i} \mathbf{W}^{t}_{p}, \quad&&\mathbf{R}^{t-1}_{i} \in \mathbb{R}^{\frac{\mathsf{p}}{\mathsf{q}} \times \frac{\mathsf{p}}{\mathsf{q}} \times \mathsf{d}}, ~\mathbf{W}^{t}_{p} \in \mathbb{R}^{\mathsf{d} \times \mathsf{d} }, \label{eq:global_query}\\
    \mathbf{f}^{t}_{i} &=  \text{Sum}\big( \mathbf{Z}^{t}_{i} \odot \mathbf{P}^{t}_{i}\big) \mathbf{W}^{t}_{f}, \quad&& \mathbf{Z}^{t}_{i} \in \mathbb{R}^{\frac{\mathsf{p}}{\mathsf{q}} \times \frac{\mathsf{p}}{\mathsf{q}} \times 1}, \mathbf{P}^{t}_{i} \in \mathbb{R}^{ \frac{\mathsf{p}}{\mathsf{q}} \times \frac{\mathsf{p}}{\mathsf{q}} \times \mathsf{d}}, \mathbf{W}^{t}_{f} \in \mathbb{R}^{\mathsf{d} \times \mathsf{d}}, \label{eq:global_out}
\end{align}
where $\mathbf{W}^{t}_{z}$, $\mathbf{W}^{t}_{p}$, and  $\mathbf{W}^{t}_{f}$ are weight matrices. $\text{Sum}(\cdot)$ denotes sum-aggregation along spatial dimensions, i.e., $\frac{\mathsf{p}}{\mathsf{q}} \times \frac{\mathsf{p}}{\mathsf{q}}$. $\text{Softmax}(\cdot)$ denotes the Softmax function that normalize scores along the same spatial dimensions.

\noindent \textbf{Objectives.} Similar to \cite{rna-seq}, we apply $\mathcal{L}_{2}$ loss to our \textit{Dual Attention} network that establishes a mapping from global feature vectors $\mathcal{F}^{G}$ and local feature cubes $\mathcal{F}^{L}$ to gene expression $\mathbf{Y}$.
We have $\mathcal{L}_{2} = \rVert \mathbf{Y} - \mathbf{M}(\mathcal{F}^{G}, \mathcal{F}^{L}) \rVert^{2}$.

\begin{table}[!t]
    \centering
    \caption{ \it Gene expression predictions. We compare with SOTA methods using the standard  Pearson Correlation Coefficient (PCC) metric. Our method consistently outperforms other methods for different gene types.}
    \vspace{3pt}
    \resizebox{\linewidth}{!}{
    \begin{tabular}{l|cccc|cc|c|cccc|cc|ccc|c}
         \toprule
         Cancer Type & \multicolumn{7}{c|}{LIHC} & \multicolumn{6}{c|}{COAD} & \multicolumn{3}{c|}{PRAD} & \multirow{2}{*}{Avg.}\\
         \cmidrule{1-17}
         Gene & CD3D & CD247 & CD3E & CD3G & CD20 & CD19 & MK167 & CD3D & CD247 & CD3E & CD3G & CD20 & CD19 & TP63 & KRT8 &KRT18 \\
         \midrule 
         HE2RNA \cite{rna-seq} &0.400&0.410&0.410&0.370&0.320&0.270&0.470&\textbf{0.430}&0.390&0.410&0.390&0.200&0.110&0.180&0.120&0.120 & 0.313\\
         ViT-S \cite{vit} &0.193&0.252&0.256&0.279&0.258&0.187&0.189&0.260&0.291&0.312&0.300&0.314&0.280&0.065&0.153&0.151&0.234\\
         ViT-MB\cite{efficientnet} &0.337&0.388&0.378&0.345&0.360&0.361&0.464&0.351&0.379&0.382&0.370&0.379&0.383&0.207&0.154&0.165&0.338\\
         CycleMLP\cite{cycle_mlp} &0.343&0.364&0.396&0.374&0.353&0.352&0.348&0.320&0.347&0.378&0.377&0.372&0.378&0.203&0.140&0.187&0.327\\ 
         MPViT\cite{mpvit}&0.365&0.358&0.377&0.350&0.379&0.356&0.491&0.311&0.361&0.371&0.352&0.374&0.413&0.205&0.142&0.199&0.338\\
         \name&\textbf{0.486}&\textbf{0.498}&\textbf{0.533}&\textbf{0.524}&\textbf{0.425}&\textbf{0.440}&\textbf{0.597}&0.415&\textbf{0.470}&\textbf{0.468}&\textbf{0.445}&\textbf{0.385}&\textbf{0.432}&\textbf{0.235}&\textbf{0.264}&\textbf{0.348}&\textbf{0.435}\\
         \bottomrule
    \end{tabular}}
    \label{tab:result}
\end{table}

\section{Experiments}
\noindent\textbf{Dataset.} We first evaluate our method on the popular dataset curated by Schmauch \textit{et al.} \cite{rna-seq}, namely TCGA dataset. It has 6 different `cancer + gene' type prediction tasks. In addition, to validate the generalization ability of our method, we directly apply our method to a clinic application, microsatellite instability (MSI) status prediction, on the three whole slide images (WSIs) dataset \cite{wsis}.
 
\noindent\textbf{Methods for Comparison.} We compare \name~framework with existing SOTA methods and possible alternatives (Tab. \ref{tab:result}). 
1) HE2RNA \cite{rna-seq}, which is the SOTA method in the gene expression prediction task.
2) ViT-S. 
We use ViT-S\footnote{Experiments show that ViT-S achieves a better performance than the original ViT-B \cite{vit}.}, which is a smaller version of ViT-B \cite{vit}.  
3) ViT-MB, CycleMLP and MPViT. Both the ViT-S and ViT-B are unable to encode the local features directly, as the number of local features is quadratically increased with respect to the number of global features. Therefore, we use MBCONV blocks, CycleMLP \cite{cycle_mlp}, and MPViT \cite{mpvit} to downsample the local features for the ViT-S;
and 4) Downsampled images. ViT architectures fail to converge if directly taking downsampled images as inputs.

\noindent\textbf{Implementation details.} Following \cite{rna-seq}, the target gene expression is log scale normalized, and we use the 5-fold cross-validation strategy as \cite{rna-seq}. Each model is trained on assigned folds before tuning on each `cancer + gene' type prediction task. Please refer to the supplementary material for more details of our model and the baselines.

\subsection{Results}
\label{sec:exp}

\noindent \textbf{Comparisons with SOTA methods.}  
We measure the Pearson Correlation Coefficient (PCC) \cite{rna-seq} between model predictions and ground truth (GT) for gene types under different cancers. Our \name~achieves the best performance across all combinations of gene types and cancer types (\cref{tab:result}). We have the following observations: 1) The low performance of ViT-S \cite{vit} indicates that only using global features ignores the fine-grained patch information; 
2) After having the local feature, the performances of ViT-MB \cite{efficientnet}, CycleMLP \cite{cycle_mlp}, and MPViT \cite{mpvit} significantly improved in PCC. The improvement demonstrates the advantages of our extracted local patch features;
and 3) Compared with the SOTA baseline HE2RNA \cite{rna-seq}, we achieve better performance, demonstrating the effectiveness of our ISG framework. 

\noindent \textbf{Comparison of Different Patch Selection Methods.} Our \name~uses \textit{Shannon Selection} module as the selector to find feature abundant patches. To demonstrate its effectiveness, we compare with commonly used patch selection methods, i.e., Image Gradient (IG), Canny Edge Detector (CANNY), DexiNed (Dex) \cite{dex}, Difference of Gaussian (DoG), Laplacian of Gaussian (LoG), and Otsu algorithm (Otsu) \cite{otsu}. They are denoted after the `-' symbol of \name.  
The results are given in \cref{tab:selection_compare}. We have the following observations: 1) our \textit{Shannon Selection} module achieves the best performance, with the averaged score at 0.435; and 2) with the same Otsu selector of HE2RNA \cite{rna-seq}, our method gets an average score of 0.380 and still outperforms HE2RNA (0.313), indicating the effectiveness of our full pipeline.

\noindent \textbf{Extra Clinic Application.} To validate the generalization ability of our method, we explore a direct clinic application: microsatellite instability (MSI) status prediction. We aim to distinguish MSI-High (MSI-H) from MSI-Stable (MSS). 
We use the standard area under the curve (AUC) \cite{rna-seq} metric. The results on the datasets provided by \cite{wsis} are given in \cref{tab:extension}.
Our \name~outperforms HE2RNA \cite{rna-seq} and achieves competitive performance compared with MSIfromHE \cite{wsis} on each set of the WSIs dataset. Note that, our \name~model is directly applied on the WSIs dataset. 

\noindent\textbf{Efficiency.} To compare the time efficiency of our model and HE2RNA, we estimate the inference time of 100 randomly sampled slide images for \name~and HE2RNA \cite{rna-seq}. For fair comparisons, we use the same GPU and slide image reading package {\tt slideio}. On average, the inference on each slide image takes 246 seconds by using our methods, while HE2RNA takes 302 seconds. 
HE2RNA squanders computations on extracting features from textureless patches, due to the Otsu selector failing to effectively filter these patches (see \cref{sec:intro}).
Furthermore, we verify the generalizability ability of our proposed \textit{Shannon Selection} Module. We apply the \textit{Shannon Selection} module in the HE2RNA for selecting patches with abundant features. This simple replacement increases the average performance of `HE2RNA-Shannon' by 7.5\%.

\begin{table}[!t]
    \begin{minipage}{\linewidth}
    \centering
    \caption{\it Comparison of using different patch selection methods.}
    \vspace{3pt}
    \resizebox{\linewidth}{!}{
    \begin{tabular}{l|cccc|cc|c|cccc|cc|ccc|c}
         \toprule
         Cancer Type & \multicolumn{7}{c|}{LIHC} & \multicolumn{6}{c|}{COAD} & \multicolumn{3}{c|}{PRAD} & \multirow{2}{*}{Avg.} \\
         \cmidrule{1-17}
         Gene & CD3D & CD247 & CD3E & CD3G & CD20 & CD19 & MK167 & CD3D & CD247 & CD3E & CD3G & CD20 & CD19 & TP63 & KRT8 &KRT18 \\
         \midrule 
         \name-IG & 0.412 & 0.428 & 0.477 & 0.485 & \textbf{0.432} & 0.426 & 0.578 & 0.428 & 0.434 & 0.457 & \textbf{0.449} & 0.378 & 0.342 & 0.210 & 0.163 & 0.277 & 0.398\\
         \name-CANNY & 0.473 & 0.445&0.511&0.484&0.373&0.396&0.583 &0.400&0.436&0.436&0.438&0.340&0.341&0.190&0.249&0.318 & 0.401\\
         \name-Dex &0.436&0.437&0.503&0.493&0.428&0.433&0.576&0.386&0.423&0.442&0.427&0.338&0.362&0.189&0.245&0.315&0.402\\
         \name-DoG &0.468&0.431&0.490&0.457&0.411&0.414&0.591&0.382&0.431&0.422&0.421&0.350&0.351&0.209&\textbf{0.273}&0.343&0.403\\
         \name-LoG &0.338&0.293&0.316&0.337&0.305&0.323&0.354&0.181&0.235&0.216&0.207&0.243&0.199&0.136&0.155&0.202&0.253\\
         \name-Otsu &0.398&0.410&0.450&0.423&0.395&0.378&0.559&0.375&0.398&0.416&0.432&0.375&0.326&0.209&0.225&0.300 & 0.380\\
         \name&\textbf{0.486}&\textbf{0.498}&\textbf{0.533}&\textbf{0.524}&0.425&\textbf{0.440}&\textbf{0.597}&\textbf{0.415}&\textbf{0.470}&\textbf{0.468}&0.445&\textbf{0.385}&\textbf{0.432}&\textbf{0.235}&0.264&\textbf{0.348} & \textbf{0.435}\\
         \bottomrule
    \end{tabular}}
    \label{tab:selection_compare}
    \vspace{5pt}
    \end{minipage}
    \begin{minipage}{0.58\linewidth}
    \centering
    \caption{ \it AUC of MSI status predictions.}
    \vspace{3pt}
    \resizebox{0.86\linewidth}{!}{
    \begin{tabular}{llccc}
         \toprule
         Dataset & \name & HE2RNA \cite{rna-seq} & MSIfromHE \cite{wsis}\\
         \midrule
         TCGA-CRC-DX &\textbf{0.86}  & 0.82 & 0.77\\
         TCGA-CRC-KR &\textbf{0.87}  & 0.83  & 0.84\\
         TCGA-STAD &0.78 & 0.76  & \textbf{0.81}\\
         \bottomrule
    \end{tabular}
    }
    \label{tab:extension}
   \end{minipage}\hfill
    \begin{minipage}{0.42\linewidth}
    \centering
    \caption{ \it Ablation study on \textit{Shannon Selection} threshold.}
    \vspace{3pt}
    \resizebox{\linewidth}{!}{
    \begin{tabular}{lccc}
         \toprule
         Threshold (bits) &$8\times10^{5}$ & $1.6\times10^{6}$&$2.4\times10^{6}$ \\
         \midrule
         Avg PCC $\uparrow$ & 0.386 & \textbf{0.390}&0.373\\
         Avg $\mathcal{L}_{2}$ $\downarrow$ &0.024&\textbf{0.023}&0.025 \\
         \bottomrule
    \end{tabular}
    }
    \label{tab:ablation_shannon}
    \end{minipage}
\end{table}

\subsection{Discussion}
\label{sec:abl} 

All experiments in this section are done by using a single model for all of the 6 prediction tasks in the TCGA dataset. Please refer to the supplementary material for more experiments and analysis.

\noindent \textbf{Shannon Selection Threshold.} 
We study the the selection threshold of the \textit{Shannon Selection} module (\cref{tab:ablation_shannon}). We achieve the best PCC with the threshold at $1.6\times10^{6}$ bits by following the mathematical implication. 
Suppose the patch texture abundance distribution follows a normal distribution $\mathcal{N}(\cdot \mid \mu,\sigma^{2})$, where $\mu$ and $\sigma^{2}$ are mean and variance. Our filter roughly keeps patches residing in the area of being at least one positive standard deviation from the mean, i.e., those patches with abundance score bigger than $\mathcal{N}(\mu + \sigma^{2} \mid \mu, \sigma^{2})$. Semantically, the patch is known to be texture abundant. A smaller selection threshold results in massive textureless patches that imposes interference signals to model inference. Reversely, a larger selection threshold excludes patches with moderate texture abundance that restricts the model from perceiving comprehensive input image features. As shown in \cref{tab:ablation_shannon}, our \name~is robust with the varying Shannon selection threshold. Moreover, even with a sub-optimal selection threshold and using the single model for all 6 prediction tasks, our framework outperforms HE2RNA in PCC (\cref{tab:result}), i.e., 0.373 vs. 0.313.

\noindent \textbf{Feature Effectiveness.} To verify the quality of extracted features by our \extractor network, we train a simple predictor, i.e., a two-layer perceptron with a ReLU activation. 
With this predictor, our \extractor network is compared to a pretrained ResNet50 (suggested by HE2RNA \cite{rna-seq}) in \cref{tab:feature}. We finetune a ResNet50 as a reference whenever possible. For the TCGA dataset \cite{rna-seq}, all patch representations are pooled to input the predictor. Using our proposed \extractor network, the predictor demonstrates a stronger PCC than the pretrained ResNet50 representations. Note that we do not present the results of a finetuned ResNet50 because of GPU memory constraints. Second, we explore the STNet dataset \cite{stnet}. It is a small-scale dataset that contains spot-level ($149 \times 149$ pixels) gene expression annotations. We use the target gene expression types selected by \cite{stnet}. Our proposed representations consistently beat both the pretrained and finetuned ResNet50 representations for the gene expression prediction task. Results demonstrate the robustness and expressiveness of our proposed \extractor network across slide image-based datasets. 

\noindent \textbf{Architectures.} We ablate the number of MBCONV blocks and frequency of using our proposed attention fusion blocks in our \textit{Dual Attention} network (\cref{tab:ablation}). Their configurations are denoted after `\name' and separated by the `-' symbol. We measure their performance with PCC and $\mathcal{L}_{2}$. Our observations are: 1) \name-10-2 tends to be an optimal setup that balances interactions between local features and global features to obtain the best PCC. However, \name-6-2 leads to the best $\mathcal{L}_{2}$. Our task emphasizes the relative changes in gene expression, thus we bias on the PCC measurement and recommend \name-10-2 as our final architecture;
and 2) There is no strict correlation between PCC and $\mathcal{L}_{2}$. The former counts an integral correlation between the prediction and the GT across all samples. The latter calculates a sample-wise deviation between the predictions and the GT. Thus, they behave differently in measuring model performance.

\begin{table}[!t]
    \begin{minipage}{\linewidth}
    \centering
    \caption{ \it Representation effectiveness evaluations. `-' denotes the result is unavailable.
    }
    \vspace{3pt}
    \resizebox{0.92\linewidth}{!}{
    \begin{tabular}{lccc}
        \toprule
         Model& ResNet50 (Pretrained) & \extractor network (Pretrained)  & ResNet50 (Finetuned)\\
         \midrule
         TCGA dataset \cite{rna-seq} &0.123 &\textbf{0.223} & - \\
         STNet dataset \cite{stnet} & 0.036 & \textbf{0.178} & 0.162\\
         \bottomrule
    \end{tabular}}
    \label{tab:feature}
    \vspace{5pt}
    \end{minipage}
    \begin{minipage}{\linewidth}
    \centering
    \caption{ \it Ablation Study on \textit{Dual Attention} network. 
    }
    \vspace{3pt}
    \resizebox{0.92\linewidth}{!}{
    \begin{tabular}{lccccccccc}
    \toprule
    Model & \name-6-2 & \name-8-2 & \name-10-2 & \name-12-2 &\name-14-2 & \name-10-1 & \name-10-2 & \name-10-3 \\
    \midrule
    Avg PCC $\uparrow$ &  0.3723&0.3750&\textbf{0.3908}&0.3601&0.3263&0.2488&\textbf{0.3908}&0.3628\\ 
    Avg $\mathcal{L}_{2}$ $\downarrow$& \textbf{0.0229}&0.00231&0.0234&0.0233&0.0239&0.0262&0.0234&0.0235\\ 
    \bottomrule
    \end{tabular}}
    \label{tab:ablation}
    \end{minipage}\hfill
\end{table}

\section{Conclusion}

In this paper, we have proposed an \name~framework to predict gene expression from histology slide images. Our key idea is to establish spatial interactions among sparsely and non-uniformly distributed feature patches of the input image for the prediction. 
To do so, we select the patches tiled at two distinct resolutions with abundant features by our \textit{Shannon Selection} module.
Then, the patches are embedded into low-dimension representations by our \extractor network trained in an unsupervised manner. Finally, we design a \textit{Dual Attention} network to refine the extracted features, to calibrate network attention on the regions of interest for gene prediction. Extensive experiments have validated the effectiveness, efficiency, and generalization ability of our method. We hope the proposed \name~framework 
can facilitate disease diagnosis and treatment development.  

\noindent \textbf{Acknowledgment.} Liyuan Pan's work was supported in part by the Beijing Institute of Technology Research Fund Program for Young Scholars.
\bibliography{egbib}

\begin{thebibliography}{41}
\providecommand{\natexlab}[1]{#1}
\providecommand{\url}[1]{\texttt{#1}}
\expandafter\ifx\csname urlstyle\endcsname\relax
  \providecommand{\doi}[1]{doi: #1}\else
  \providecommand{\doi}{doi: \begingroup \urlstyle{rm}\Url}\fi

\bibitem[Chen et~al.(2021)Chen, Xie, Ge, Liang, and Luo]{cycle_mlp}
Shoufa Chen, Enze Xie, Chongjian Ge, Ding Liang, and Ping Luo.
\newblock Cyclemlp: {A} mlp-like architecture for dense prediction.
\newblock \emph{CoRR}, abs/2107.10224, 2021.
\newblock URL \url{https://arxiv.org/abs/2107.10224}.

\bibitem[Chen and He(2021)]{sia2}
Xinlei Chen and Kaiming He.
\newblock Exploring simple siamese representation learning.
\newblock In \emph{{IEEE} Conference on Computer Vision and Pattern
  Recognition, {CVPR} 2021, virtual, June 19-25, 2021}, pages 15750--15758.
  Computer Vision Foundation / {IEEE}, 2021.
\newblock URL
  \url{https://openaccess.thecvf.com/content/CVPR2021/html/Chen\_Exploring\_Simple\_Siamese\_Representation\_Learning\_CVPR\_2021\_paper.html}.

\bibitem[Coudray et~al.(2018)Coudray, Ocampo, Sakellaropoulos, Narula, Snuderl,
  Fenyö, Moreira, Razavian, and Tsirigos]{m1}
Nicolas Coudray, Paolo Ocampo, Theodore Sakellaropoulos, Navneet Narula, Matija
  Snuderl, David Fenyö, Andre Moreira, Narges Razavian, and Aristotelis
  Tsirigos.
\newblock Classification and mutation prediction from non–small cell lung
  cancer histopathology images using deep learning.
\newblock \emph{Nature Medicine}, 24, 10 2018.
\newblock \doi{10.1038/s41591-018-0177-5}.

\bibitem[Dawood et~al.(2021)Dawood, Branson, Rajpoot, and
  Minhas]{stain_learning}
Muhammad Dawood, Kim Branson, Nasir Rajpoot, and Fayyaz ul Amir~Afsar Minhas.
\newblock All you need is color: Image based spatial gene expression prediction
  using neural stain learning.
\newblock 08 2021.

\bibitem[Dosovitskiy et~al.(2021)Dosovitskiy, Beyer, Kolesnikov, Weissenborn,
  Zhai, Unterthiner, Dehghani, Minderer, Heigold, Gelly, Uszkoreit, and
  Houlsby]{vit}
Alexey Dosovitskiy, Lucas Beyer, Alexander Kolesnikov, Dirk Weissenborn,
  Xiaohua Zhai, Thomas Unterthiner, Mostafa Dehghani, Matthias Minderer, Georg
  Heigold, Sylvain Gelly, Jakob Uszkoreit, and Neil Houlsby.
\newblock An image is worth 16x16 words: Transformers for image recognition at
  scale.
\newblock In \emph{9th International Conference on Learning Representations,
  {ICLR} 2021, Virtual Event, Austria, May 3-7, 2021}. OpenReview.net, 2021.
\newblock URL \url{https://openreview.net/forum?id=YicbFdNTTy}.

\bibitem[He et~al.(2020)He, Bergenstråhle, Stenbeck, Abid, Andersson, Borg,
  Maaskola, Lundeberg, and Zou]{stnet}
Bryan He, Ludvig Bergenstråhle, Linnea Stenbeck, Abubakar Abid, Alma
  Andersson, Ake Borg, Jonas Maaskola, Joakim Lundeberg, and James Zou.
\newblock Integrating spatial gene expression and breast tumour morphology via
  deep learning.
\newblock \emph{Nature Biomedical Engineering}, 4:\penalty0 1--8, 08 2020.
\newblock \doi{10.1038/s41551-020-0578-x}.

\bibitem[He et~al.(2016)He, Zhang, Ren, and Sun]{resnet}
Kaiming He, Xiangyu Zhang, Shaoqing Ren, and Jian Sun.
\newblock Deep residual learning for image recognition.
\newblock In \emph{2016 {IEEE} Conference on Computer Vision and Pattern
  Recognition, {CVPR} 2016, Las Vegas, NV, USA, June 27-30, 2016}, pages
  770--778. {IEEE} Computer Society, 2016.
\newblock \doi{10.1109/CVPR.2016.90}.
\newblock URL \url{https://doi.org/10.1109/CVPR.2016.90}.

\bibitem[Hou et~al.(2016)Hou, Samaras, Kur{\c{c}}, Gao, Davis, and Saltz]{c2}
Le~Hou, Dimitris Samaras, Tahsin~M. Kur{\c{c}}, Yi~Gao, James~E. Davis, and
  Joel~H. Saltz.
\newblock Patch-based convolutional neural network for whole slide tissue image
  classification.
\newblock In \emph{2016 {IEEE} Conference on Computer Vision and Pattern
  Recognition, {CVPR} 2016, Las Vegas, NV, USA, June 27-30, 2016}, pages
  2424--2433. {IEEE} Computer Society, 2016.
\newblock \doi{10.1109/CVPR.2016.266}.
\newblock URL \url{https://doi.org/10.1109/CVPR.2016.266}.

\bibitem[Hutter(2012)]{uai}
Marcus Hutter.
\newblock Universal artificial intelligence: Sequential decisions based on
  algorithmic probability.
\newblock 04 2012.
\newblock \doi{10.1007/b138233}.

\bibitem[Karras et~al.(2020)Karras, Laine, Aittala, Hellsten, Lehtinen, and
  Aila]{stylegan2}
Tero Karras, Samuli Laine, Miika Aittala, Janne Hellsten, Jaakko Lehtinen, and
  Timo Aila.
\newblock Analyzing and improving the image quality of stylegan.
\newblock In \emph{2020 {IEEE/CVF} Conference on Computer Vision and Pattern
  Recognition, {CVPR} 2020, Seattle, WA, USA, June 13-19, 2020}, pages
  8107--8116. Computer Vision Foundation / {IEEE}, 2020.
\newblock \doi{10.1109/CVPR42600.2020.00813}.
\newblock URL
  \url{https://openaccess.thecvf.com/content\_CVPR\_2020/html/Karras\_Analyzing\_and\_Improving\_the\_Image\_Quality\_of\_StyleGAN\_CVPR\_2020\_paper.html}.

\bibitem[Karras et~al.(2021)Karras, Aittala, Laine, H{\"{a}}rk{\"{o}}nen,
  Hellsten, Lehtinen, and Aila]{stylegan3}
Tero Karras, Miika Aittala, Samuli Laine, Erik H{\"{a}}rk{\"{o}}nen, Janne
  Hellsten, Jaakko Lehtinen, and Timo Aila.
\newblock Alias-free generative adversarial networks.
\newblock \emph{CoRR}, abs/2106.12423, 2021.
\newblock URL \url{https://arxiv.org/abs/2106.12423}.

\bibitem[Kather et~al.(2019)Kather, Pearson, Halama, Jäger, Krause, Loosen,
  Marx, Boor, Tacke, Neumann, Grabsch, Yoshikawa, Brenner, Chang-Claude,
  Hoffmeister, Trautwein, and Luedde]{wsis}
Jakob Kather, Alexander Pearson, Niels Halama, Dirk Jäger, Jeremias Krause,
  Sven Loosen, Alexander Marx, Peter Boor, Frank Tacke, Ulf Neumann, Heike
  Grabsch, Takaki Yoshikawa, Hermann Brenner, Jenny Chang-Claude, Michael
  Hoffmeister, Christian Trautwein, and Tom Luedde.
\newblock Deep learning can predict microsatellite instability directly from
  histology in gastrointestinal cancer.
\newblock \emph{Nature Medicine}, 25, 07 2019.
\newblock \doi{10.1038/s41591-019-0462-y}.

\bibitem[Krizhevsky et~al.(2017)Krizhevsky, Sutskever, and Hinton]{alexnet}
Alex Krizhevsky, Ilya Sutskever, and Geoffrey~E. Hinton.
\newblock Imagenet classification with deep convolutional neural networks.
\newblock \emph{Commun. {ACM}}, 60\penalty0 (6):\penalty0 84--90, 2017.
\newblock \doi{10.1145/3065386}.
\newblock URL \url{http://doi.acm.org/10.1145/3065386}.

\bibitem[Kubiak et~al.(2021)Kubiak, Mustafa, Phillipson, Jolly, and
  Hadfield]{utl1}
Nikolina Kubiak, Armin Mustafa, Graeme Phillipson, Stephen Jolly, and Simon
  Hadfield.
\newblock {SILT:} self-supervised lighting transfer using implicit image
  decomposition.
\newblock \emph{CoRR}, abs/2110.12914, 2021.
\newblock URL \url{https://arxiv.org/abs/2110.12914}.

\bibitem[Kulikov and Lempitsky(2020)]{seg2}
Victor Kulikov and Victor~S. Lempitsky.
\newblock Instance segmentation of biological images using harmonic embeddings.
\newblock In \emph{2020 {IEEE/CVF} Conference on Computer Vision and Pattern
  Recognition, {CVPR} 2020, Seattle, WA, USA, June 13-19, 2020}, pages
  3842--3850. Computer Vision Foundation / {IEEE}, 2020.
\newblock \doi{10.1109/CVPR42600.2020.00390}.
\newblock URL
  \url{https://openaccess.thecvf.com/content\_CVPR\_2020/html/Kulikov\_Instance\_Segmentation\_of\_Biological\_Images\_Using\_Harmonic\_Embeddings\_CVPR\_2020\_paper.html}.

\bibitem[Lee et~al.(2021)Lee, Kim, Willette, and Hwang]{mpvit}
Youngwan Lee, Jonghee Kim, Jeffrey Willette, and Sung~Ju Hwang.
\newblock Mpvit: Multi-path vision transformer for dense prediction.
\newblock \emph{CoRR}, abs/2112.11010, 2021.
\newblock URL \url{https://arxiv.org/abs/2112.11010}.

\bibitem[Liu et~al.(2021)Liu, Neophytou, Sengupta, and Sommerlade]{sia1}
Yang Liu, Alexandros Neophytou, Sunando Sengupta, and Eric Sommerlade.
\newblock Relighting images in the wild with a self-supervised siamese
  auto-encoder.
\newblock In \emph{{IEEE} Winter Conference on Applications of Computer Vision,
  {WACV} 2021, Waikoloa, HI, USA, January 3-8, 2021}, pages 32--40. {IEEE},
  2021.
\newblock \doi{10.1109/WACV48630.2021.00008}.
\newblock URL \url{https://doi.org/10.1109/WACV48630.2021.00008}.

\bibitem[Ma et~al.(2019)Ma, Liu, Zhang, Ye, Yin, and Johnson]{remote}
Lei Ma, Yu~Liu, Xueliang Zhang, Yuanxin Ye, Gaofei Yin, and Brian Johnson.
\newblock Deep learning in remote sensing applications: A meta-analysis and
  review.
\newblock \emph{ISPRS Journal of Photogrammetry and Remote Sensing},
  152:\penalty0 166--177, 04 2019.
\newblock \doi{10.1016/j.isprsjprs.2019.04.015}.

\bibitem[MacKay(2003)]{entropy}
David J.~C. MacKay.
\newblock \emph{Information theory, inference, and learning algorithms}.
\newblock Cambridge University Press, 2003.
\newblock ISBN 978-0-521-64298-9.

\bibitem[Madaan et~al.(2021)Madaan, Yoon, Li, Liu, and Hwang]{utl2}
Divyam Madaan, Jaehong Yoon, Yuanchun Li, Yunxin Liu, and Sung~Ju Hwang.
\newblock Rethinking the representational continuity: Towards unsupervised
  continual learning.
\newblock \emph{CoRR}, abs/2110.06976, 2021.
\newblock URL \url{https://arxiv.org/abs/2110.06976}.

\bibitem[Meng et~al.(2021)Meng, Zhang, Gao, Zhao, Yang, Qian, Huang, and
  Zheng]{seg1}
Yanda Meng, Hongrun Zhang, Dongxu Gao, Yitian Zhao, Xiaoyun Yang, Xuesheng
  Qian, Xiaowei Huang, and Yalin Zheng.
\newblock {BI-GCN:} boundary-aware input-dependent graph convolution network
  for biomedical image segmentation.
\newblock \emph{CoRR}, abs/2110.14775, 2021.
\newblock URL \url{https://arxiv.org/abs/2110.14775}.

\bibitem[Otsu(1979)]{otsu}
Nobuyuki Otsu.
\newblock A threshold selection method from gray-level histograms.
\newblock \emph{Systems, Man and Cybernetics, IEEE Transactions on},
  9:\penalty0 62--66, 01 1979.

\bibitem[Perez et~al.(2019)Perez, Avila, and Valle]{c1}
F{\'{a}}bio Perez, Sandra Avila, and Eduardo Valle.
\newblock Solo or ensemble? choosing a {CNN} architecture for melanoma
  classification.
\newblock In \emph{{IEEE} Conference on Computer Vision and Pattern Recognition
  Workshops, {CVPR} Workshops 2019, Long Beach, CA, USA, June 16-20, 2019},
  pages 2775--2783. Computer Vision Foundation / {IEEE}, 2019.
\newblock \doi{10.1109/CVPRW.2019.00336}.
\newblock URL
  \url{http://openaccess.thecvf.com/content\_CVPRW\_2019/html/ISIC/Perez\_Solo\_or\_Ensemble\_Choosing\_a\_CNN\_Architecture\_for\_Melanoma\_Classification\_CVPRW\_2019\_paper.html}.

\bibitem[Poland and Hutter(2005)]{mdl}
Jan Poland and Marcus Hutter.
\newblock Asymptotics of discrete mdl for online prediction.
\newblock \emph{Information Theory, IEEE Transactions on}, 51:\penalty0 3780 --
  3795, 12 2005.
\newblock \doi{10.1109/TIT.2005.856956}.

\bibitem[Poma et~al.(2021)Poma, Sappa, Humanante, and Akbarinia]{dex}
Xavier~Soria Poma, {\'{A}}ngel~D. Sappa, Patricio Humanante, and Arash
  Akbarinia.
\newblock Dense extreme inception network for edge detection.
\newblock \emph{CoRR}, abs/2112.02250, 2021.
\newblock URL \url{https://arxiv.org/abs/2112.02250}.

\bibitem[Richardson et~al.(2021)Richardson, Alaluf, Patashnik, Nitzan, Azar,
  Shapiro, and Cohen{-}Or]{psp}
Elad Richardson, Yuval Alaluf, Or~Patashnik, Yotam Nitzan, Yaniv Azar, Stav
  Shapiro, and Daniel Cohen{-}Or.
\newblock Encoding in style: {A} stylegan encoder for image-to-image
  translation.
\newblock In \emph{{IEEE} Conference on Computer Vision and Pattern
  Recognition, {CVPR} 2021, virtual, June 19-25, 2021}, pages 2287--2296.
  Computer Vision Foundation / {IEEE}, 2021.
\newblock URL
  \url{https://openaccess.thecvf.com/content/CVPR2021/html/Richardson\_Encoding\_in\_Style\_A\_StyleGAN\_Encoder\_for\_Image-to-Image\_Translation\_CVPR\_2021\_paper.html}.

\bibitem[Schaumberg et~al.(2018)Schaumberg, Rubin, and Fuchs]{m2}
Andrew Schaumberg, Mark Rubin, and Thomas Fuchs.
\newblock H\&e-stained whole slide image deep learning predicts spop mutation
  state in prostate cancer.
\newblock 10 2018.
\newblock \doi{10.1101/064279}.

\bibitem[Schmauch et~al.(2020)Schmauch, Romagnoni, Pronier, Saillard, Maillé,
  Calderaro, Kamoun, Sefta, Toldo, Zaslavskiy, Clozel, Moarii, Courtiol, and
  Wainrib]{rna-seq}
Benoit Schmauch, Alberto Romagnoni, Elodie Pronier, Charlie Saillard, Pascale
  Maillé, Julien Calderaro, Aurélie Kamoun, Meriem Sefta, Sylvain Toldo,
  Mikhail Zaslavskiy, Thomas Clozel, Matahi Moarii, Pierre Courtiol, and Gilles
  Wainrib.
\newblock A deep learning model to predict rna-seq expression of tumours from
  whole slide images.
\newblock \emph{Nature Communications}, 11, 08 2020.
\newblock \doi{10.1038/s41467-020-17678-4}.

\bibitem[Scott(2011)]{edge}
E.U. Scott.
\newblock Digital image processing and analysis: human and computer vision
  applications with cviptools.
\newblock \emph{Digital Image Processing and Analysis: Human and Computer
  Vision Applications with CVIP Tools}, 01 2011.

\bibitem[Ståhl et~al.(2016)Ståhl, Salmén, Vickovic, Lundmark,
  Fernandez~Navarro, Magnusson, Giacomello, Asp, Westholm, Huss, Mollbrink,
  Linnarsson, Codeluppi, Borg, Pontén, Costea, Sahlén, Mulder, Bergmann, and
  Frisén]{spatial}
Patrik Ståhl, Fredrik Salmén, Sanja Vickovic, Anna Lundmark, Jose
  Fernandez~Navarro, Jens Magnusson, Stefania Giacomello, Michaela Asp, Jakub
  Westholm, Mikael Huss, Annelie Mollbrink, Sten Linnarsson, Simone Codeluppi,
  Åke Borg, Fredrik Pontén, Paul Costea, Pelin~Akan Sahlén, Jan Mulder, Olaf
  Bergmann, and Jonas Frisén.
\newblock Visualization and analysis of gene expression in tissue sections by
  spatial transcriptomics.
\newblock \emph{Science}, 353:\penalty0 78--82, 07 2016.
\newblock \doi{10.1126/science.aaf2403}.

\bibitem[Tan and Le(2019)]{efficientnet}
Mingxing Tan and Quoc~V. Le.
\newblock Efficientnet: Rethinking model scaling for convolutional neural
  networks.
\newblock In Kamalika Chaudhuri and Ruslan Salakhutdinov, editors,
  \emph{Proceedings of the 36th International Conference on Machine Learning,
  {ICML} 2019, 9-15 June 2019, Long Beach, California, {USA}}, volume~97 of
  \emph{Proceedings of Machine Learning Research}, pages 6105--6114. {PMLR},
  2019.
\newblock URL \url{http://proceedings.mlr.press/v97/tan19a.html}.

\bibitem[Vaswani et~al.(2017{\natexlab{a}})Vaswani, Shazeer, Parmar, Uszkoreit,
  Jones, Gomez, Kaiser, and Polosukhin]{att}
Ashish Vaswani, Noam Shazeer, Niki Parmar, Jakob Uszkoreit, Llion Jones, Aidan
  Gomez, Lukasz Kaiser, and Illia Polosukhin.
\newblock Attention is all you need.
\newblock 06 2017{\natexlab{a}}.

\bibitem[Vaswani et~al.(2017{\natexlab{b}})Vaswani, Shazeer, Parmar, Uszkoreit,
  Jones, Gomez, Kaiser, and Polosukhin]{transformer}
Ashish Vaswani, Noam Shazeer, Niki Parmar, Jakob Uszkoreit, Llion Jones,
  Aidan~N. Gomez, Lukasz Kaiser, and Illia Polosukhin.
\newblock Attention is all you need.
\newblock In Isabelle Guyon, Ulrike von Luxburg, Samy Bengio, Hanna~M. Wallach,
  Rob Fergus, S.~V.~N. Vishwanathan, and Roman Garnett, editors, \emph{Advances
  in Neural Information Processing Systems 30: Annual Conference on Neural
  Information Processing Systems 2017, December 4-9, 2017, Long Beach, CA,
  {USA}}, pages 5998--6008, 2017{\natexlab{b}}.
\newblock URL
  \url{https://proceedings.neurips.cc/paper/2017/hash/3f5ee243547dee91fbd053c1c4a845aa-Abstract.html}.

\bibitem[Wang et~al.(2008)Wang, Gerstein, and Snyder]{bulk}
Zhong Wang, Mark Gerstein, and Michael Snyder.
\newblock Rna-seq: A revolutionary tool for transcriptomics.
\newblock \emph{Nature reviews. Genetics}, 10:\penalty0 57--63, 12 2008.
\newblock \doi{10.1038/nrg2484}.

\bibitem[White and Rosenblatt(1963)]{mlp}
B.~White and Frank Rosenblatt.
\newblock Principles of neurodynamics: Perceptrons and the theory of brain
  mechanisms.
\newblock \emph{The American Journal of Psychology}, 76:\penalty0 705, 12 1963.
\newblock \doi{10.2307/1419730}.

\bibitem[Wu et~al.(2021)Wu, Lischinski, and Shechtman]{stylespace}
Zongze Wu, Dani Lischinski, and Eli Shechtman.
\newblock Stylespace analysis: Disentangled controls for stylegan image
  generation.
\newblock In \emph{{IEEE} Conference on Computer Vision and Pattern
  Recognition, {CVPR} 2021, virtual, June 19-25, 2021}, pages 12863--12872.
  Computer Vision Foundation / {IEEE}, 2021.
\newblock URL
  \url{https://openaccess.thecvf.com/content/CVPR2021/html/Wu\_StyleSpace\_Analysis\_Disentangled\_Controls\_for\_StyleGAN\_Image\_Generation\_CVPR\_2021\_paper.html}.

\bibitem[Yang et~al.(2022)Yang, Hossain, Gedeon, and Rahman]{s2f}
Yan Yang, Md.~Zakir Hossain, Tom Gedeon, and Shafin Rahman.
\newblock {S2FGAN:} semantically aware interactive sketch-to-face translation.
\newblock In \emph{{IEEE/CVF} Winter Conference on Applications of Computer
  Vision, {WACV} 2022, Waikoloa, HI, USA, January 3-8, 2022}, pages 3162--3171.
  {IEEE}, 2022.
\newblock \doi{10.1109/WACV51458.2022.00322}.
\newblock URL \url{https://doi.org/10.1109/WACV51458.2022.00322}.

\bibitem[Yue et~al.(2021)Yue, Sun, Kuang, Wei, Torr, Zhang, and Lin]{vitpd}
Xiaoyu Yue, Shuyang Sun, Zhanghui Kuang, Meng Wei, Philip Torr, Wayne Zhang,
  and Dahua Lin.
\newblock Vision transformer with progressive sampling.
\newblock pages 377--386, 10 2021.
\newblock \doi{10.1109/ICCV48922.2021.00044}.

\bibitem[Zbontar et~al.(2021)Zbontar, Jing, Misra, LeCun, and Deny]{cor1}
Jure Zbontar, Li~Jing, Ishan Misra, Yann LeCun, and St{\'{e}}phane Deny.
\newblock Barlow twins: Self-supervised learning via redundancy reduction.
\newblock In Marina Meila and Tong Zhang, editors, \emph{Proceedings of the
  38th International Conference on Machine Learning, {ICML} 2021, 18-24 July
  2021, Virtual Event}, volume 139 of \emph{Proceedings of Machine Learning
  Research}, pages 12310--12320. {PMLR}, 2021.
\newblock URL \url{http://proceedings.mlr.press/v139/zbontar21a.html}.

\bibitem[Zeng et~al.(2022)Zeng, Wei, Yu, Yin, Yuan, Li, Tang, Lu, and
  Yang]{st_transformer}
Yuansong Zeng, Zhuoyi Wei, Weijiang Yu, Rui Yin, Yuchen Yuan, Bingling Li,
  Zhonghui Tang, Yutong Lu, and Yuedong Yang.
\newblock Spatial transcriptomics prediction from histology jointly through
  transformer and graph neural networks.
\newblock \emph{Briefings in Bioinformatics}, 07 2022.
\newblock \doi{10.1093/bib/bbac297}.

\bibitem[Zhang et~al.(2018)Zhang, Isola, Efros, Shechtman, and Wang]{lpips}
Richard Zhang, Phillip Isola, Alexei~A. Efros, Eli Shechtman, and Oliver Wang.
\newblock The unreasonable effectiveness of deep features as a perceptual
  metric.
\newblock In \emph{2018 {IEEE} Conference on Computer Vision and Pattern
  Recognition, {CVPR} 2018, Salt Lake City, UT, USA, June 18-22, 2018}, pages
  586--595. Computer Vision Foundation / {IEEE} Computer Society, 2018.
\newblock \doi{10.1109/CVPR.2018.00068}.
\newblock URL
  \url{http://openaccess.thecvf.com/content\_cvpr\_2018/html/Zhang\_The\_Unreasonable\_Effectiveness\_CVPR\_2018\_paper.html}.

\end{thebibliography}

\end{document}